\documentclass[letterpaper, 10 pt, conference]{ieeeconf}  

\IEEEoverridecommandlockouts                              

\usepackage{graphicx}
\usepackage{booktabs}
\usepackage{amsmath}
\usepackage[noadjust]{cite}
\usepackage{textcomp}
\usepackage{amsfonts}
\usepackage{bm}
\usepackage{multirow}
\usepackage{graphicx}
\usepackage{subcaption}
\usepackage{color}
\usepackage[normalem]{ulem}

\title{\LARGE \bf
Adaptive Environment-Aware Robotic Arm Reaching Based on a Bio-Inspired Neurodynamical Computational Framework
}

\author{Dimitrios Chatziparaschis,$^{1}$ Shan Zhong,$^{2}$ Vasileios Christopoulos,$^{2,3}$ and Konstantinos Karydis$^{1}$
\thanks{$^{1}$ ~Dept. of Electrical and Computer Engineering, ~$^{2}$ Neuroscience graduate program, ~$^{3}$~Dept. of Bioengineering, Univ. of California, Riverside, 900 University Avenue, Riverside, CA 92521, USA;
{\tt\footnotesize\{dchat013, szhon028, christov, karydis\}@ucr.edu}.}%
\thanks{We gratefully acknowledge the support of NSF \# CMMI-2133084, ONR \# N00014-19-1-2264 and NINDS \# U01NS132788. 
Any opinions, findings, and conclusions or recommendations expressed in this material are those of the authors and do not necessarily reflect the views of the funding agencies.}
}

\begin{document}

\maketitle
\thispagestyle{empty}
\pagestyle{empty}

\begin{abstract}

Bio-inspired robotic systems are capable of adaptive learning, scalable control, and efficient information processing. 
Enabling real-time decision-making for such systems is critical to respond to dynamic changes in the environment. 
We focus on dynamic target tracking in open areas using a robotic six-degree-of-freedom manipulator with a bird-eye view camera for visual feedback, and by deploying the Neurodynamical Computational Framework (NeuCF). 
NeuCF is a recently developed bio-inspired model for target tracking based on Dynamic Neural Fields (DNFs) and Stochastic Optimal Control (SOC) theory. 
It has been trained for reaching actions on a planar surface toward localized visual beacons, and it can re-target or generate stop signals on the fly based on changes in the environment (e.g., a new target has emerged, or an existing one has been removed). 
We evaluated our system over various target-reaching scenarios. 
In all experiments, NeuCF had high end-effector positional accuracy, generated smooth trajectories, and provided reduced path lengths compared with a baseline cubic polynomial trajectory generator. 
In all, the developed system offers a robust and dynamic-aware robotic manipulation approach that affords real-time decision-making.

\end{abstract}


\section{Introduction}

Object manipulation is fundamental for the integration of robots into a world designed around humans. 
Several distinctive robotic manipulation works have been developed over the years employing humanoids (e.g.,~\cite{li2018rlmanipulationhumanoid}), quadrupeds (e.g.,~\cite{zimmermann2021spotminigrasp}), and soft robotics (e.g.,~\cite{shi2022multifinger}). 
To perform manipulation tasks effectively, robots must be able to understand their surrounding environment, localize available objects, and plan and execute accurate and adaptive reaching actions.

A growing number of studies seek to employ bio-inspiration for robotic manipulation~\cite{mandlekar2020irisofflinemanipulationdata,katyara2021humaninspiredrobotmanipulation,hueang2022prostheticbiosignals,ciocarlie2009hand}. 
Such methods aim to map the neural mechanisms underlying decision-making and motor control in humans and animals to support robot cognition~\cite{erlhagen2006DNFcognitiverobot}. 
In turn, human-like robot motion may also help provide a deeper understanding of the nature of human motor behavior~\cite{hersch2006model, hyondong2017bioinspiredmultirobots}. Thus, exploring bio-inspired methodologies and learning behaviors is important to develop and attain cognitive abilities in complex robotic applications.

Bio-inspired methodologies employed in robotics can be broadly split into three main categories; Spiking Neural Networks (SNNs), Liquid State Machines (LSMs), and Dynamic Neural Fields (DNFs). 
SNNs are activated by discrete events called spikes or action potentials~\cite{dayan2005neurosiencetheoretical}. 
SNNs have been successfully deployed in 3D path planning~\cite{steffen2020snn3dplanning} and robotic manipulator motion learning~\cite{camilo2018ssnmotorreaching}. 
They have served also as internal models in reinforcement-learning-based target reaching training~\cite{vasquez2019ssnrlreaching}, and in fault-tolerant flight control~\cite{weiSNNflightcontroller_faulttolerant}.  
However, their training complexity may limit their applicability for real-time and efficient decision-making, which is crucial in modern robotic applications. 
To this end, LSMs can serve as a basis for more dynamic-aware bio-inspired methods. LSMs, based on reservoir computing~\cite{tanaka2019reservoir}, consist of a large number of recurrently connected neurons used to perform computations and recognize input patterns~\cite{maass2002lsmbook} and have been used in robotic manipulation tasks~\cite{alberto2017positioninglsm,de2017robotarmlsm}. 

\begin{figure*}
\vspace{6pt}
    \centering
    \includegraphics[height=7cm]{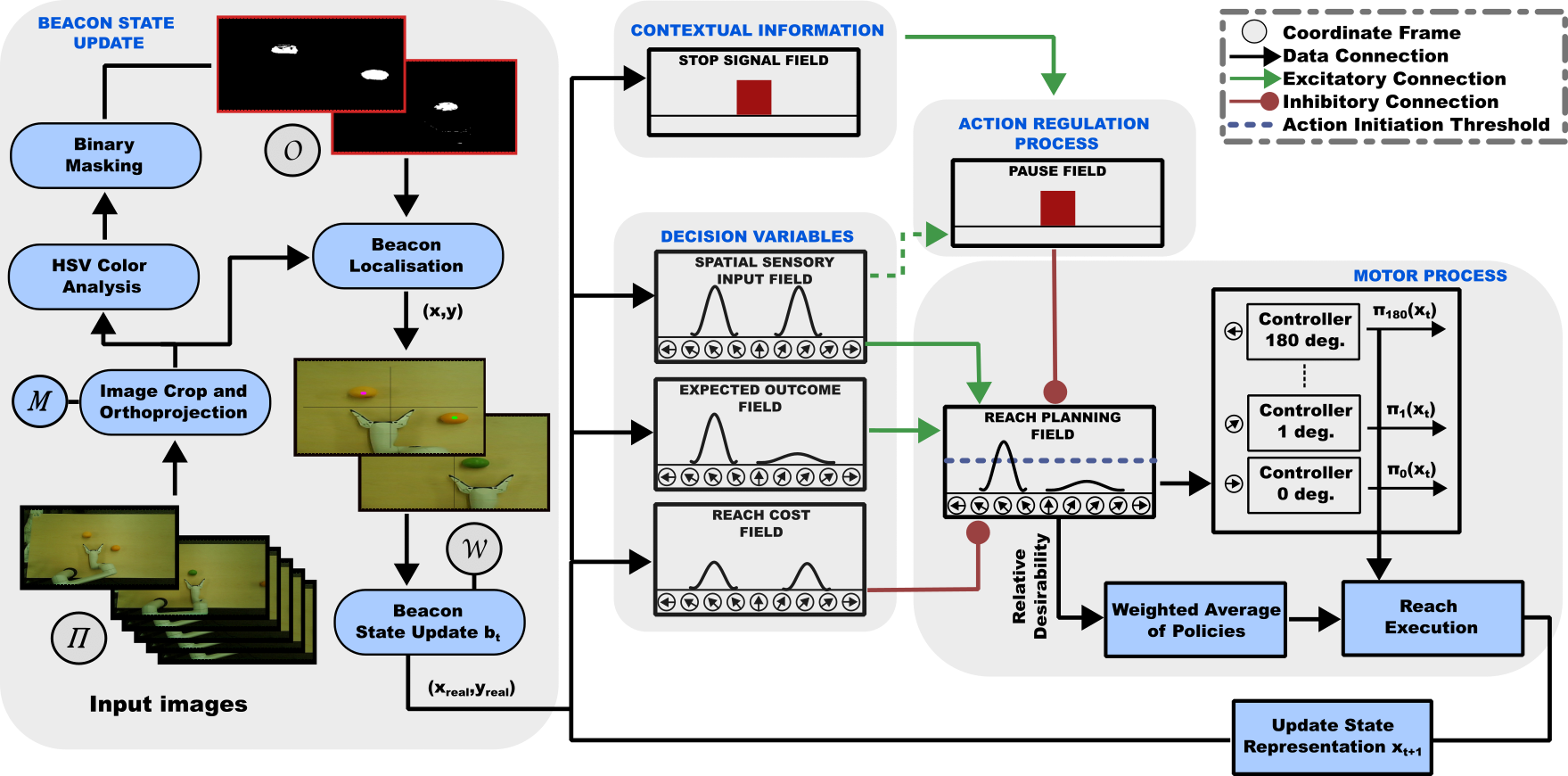}
    \caption{Overall System Architecture. The system receives images to compute beacon locations. The reach planning field encodes the intended movement direction by combining inputs from disparate sources. The relative desirability value for each action policy captures its attractiveness compared to alternatives and acts as a weight to compute the final action policy.}
    \label{fig:system}
    \vspace{-15pt}
\end{figure*}

DNFs have emerged as a prominent theory for real-time decision-making~\cite{Sandamirskaya_2014}, and have been used to model aspects of perception~\cite{faubel2008learning, erlhagen2006DNFcognitiverobot} and action~\cite{katerishich2023dnfautonomousnavigation, cunha2020humanrobotaction, malheiro2017frameworkDNFvrep, erlhagen2002dynamic} in both biological and non-biological systems. 
Several studies have applied the DNF theory in robotic reaching and grasping. 
Specifically, Ferreira $et~al.$~\cite{ferreira2021rapidlearningdnf} presented a time-aware DNF-based control architecture for human-robot interaction. 
Strauss $et~al.$~\cite{strauss2015choice} developed a DNF-based architecture for integrating visual input, decision-making, and motor planning in a robotic reaching task, whereas Knips $et~al.$~\cite{Knips_2017} proposed a neural dynamic architecture that utilizes DNFs for perception, movement generation, and action selection, enabling online updating of motor plans based on sensory feedback. 
While these approaches have made significant contributions to the field, they do not explicitly address the integration of optimal control principles for generating efficient reaching trajectories in the presence of multiple competing targets and uncertain states. 
Also, they do not provide a unified framework that can emulate a variety of motor tasks in dynamic environments, like deciding between alternative options, stopping unwanted or inappropriate ongoing actions, and switching actions in response to environmental changes. 

To address these limitations, we deploy the Neurodynamical Computational Framework (NeuCF), which merges DNF and Stochastic Optimal Control (SOC) theory to generate adaptive reaching trajectories in a neurally plausible manner, integrating action selection, stopping, and switching mechanisms. NeuCF was designed and developed based on recent studies that modeled motor and neural basis in diverse motor decisions performed by human subjects in dynamic and uncertain environments~\cite{christopoulos2015probability, christopoulos2015biologically, zhong2022neurocomputational}. 
The architecture and parameters of NeuCF have been tuned based on reaching movements performed by human subjects in dynamic environments, ensuring biologically plausible behavior. 
In this work, we demonstrate the potential of the NeuCF controller for accurate and dynamic environment-aware target reaching in robotic manipulation. 
We evaluate NeuCF's effectiveness in enabling a six-degree-of-freedom manipulator with a bird-eye view camera for target tracking to perform reaching tasks in dynamic environments with multiple competing targets.

\section{Materials and Methods}\label{seq:systemsetup}

\subsection{Neurodynamical Computational Framework (NeuCF)}
In earlier work, we have proposed a framework for action regulation tasks involving motor inhibition, including selecting, stopping, and switching of actions~\cite{christopoulos2015biologically,zhong2022neurocomputational,zhong2023computational}. 
The framework combines DNFs, which simulate the evolution of neural activity over a continuous space where neurons interact through excitatory and inhibitory connections based on their spatial proximity in the field~\cite{erlhagen2002dynamic}, and SOC theory, which seeks to find the optimal control policy that minimizes a cost function in a stochastic system. 
The network of DNFs represents the neural circuitry responsible for various aspects of action regulation (e.g., sensory input, expected outcome, cost, context signal, action planning, and action execution) while enabling the representation and selection of multiple potential targets. 
Optimal reaching strategies are created using SOC theory, based on the current environment state, and expected rewards and costs linked with each action. 

Figure~\ref{fig:system} depicts the overall model architecture of our system. 
The neurodynamical framework receives image inputs from the camera to compute beacon locations.  
Within the neurodynamical framework, the ``Reach Planning'' field purpose is twofold: 1) initiation of the stochastic optimal controllers generating action plans along specific directions, 2) integration of diverse information sources related to actions, objectives, and situational requirements into a single measure representing action desirability or ``attractiveness.'' 
Excitatory input from the ``spatial sensory input'' field (denoted by green arrows) informs the reach planning field about the angular positions of targets in an egocentric reference frame and the ``expected outcome'' field conveys expected rewards linked to motion in specific directions. 
Inhibitory feedback (denoted by red arrows) is also received from the ``reach cost'' field, quantifying the ``cost'' needed to move in designated directions, and the ``pause'' field, which serves as a rapid suppressor of planned or ongoing actions during necessary action inhibition. 
The activity of the reach planning field is the sum of the outputs of these fields, corrupted by Gaussian noise. 
This activity is then normalized to represent each neuron's relative desirability against other available options at any particular time and state, with elevated neuronal activity signifying increased desirability to move in its favored direction. 
Neurons in the reach planning field are linked to an optimal control scheme responsible for generating reaching actions. 
When a neuron's activity exceeds a preset ``action initiation threshold,'' the associated controller is activated, generating an optimal policy $\pi_i$ (i.e. a series of motor commands) directed toward that neuron's preferred direction. 

The optimal policy $\pi_j$ is given by minimizing 
%
%
\begin{multline}
     J_j(\bm{x}_t,\bm{\pi}_j) = (\bm{x}_{T_j}-S\bm{p}_j)^TQ_{T_j}(\bm{x}_{T_j}-S\bm{p}_j)\\
     +\sum_{t=1}^{T_j-1}\bm{\pi}_j(\bm{x}_t)^TR\bm{\pi}_j(\bm{x}_t)\;,
\end{multline}

\noindent where $\bm{\pi}_j(\bm{x}_t)$ represents the policy with $t\in[1,T_J]$ for reaching in the preferred direction $\varphi_j$ and $T_j$ denotes the time taken to reach the position $\bm{p_j}$ = [r\;cos($\varphi_j$), r\;sin($\varphi_j$)]. 
Variable $r$ represents the distance between the current arm position and the target location encoded by the neuron \textit{j}. 
Vector $\bm{x}_{T_j}$ represents the state when the reaching movement concludes, and matrix $S$ extracts the actual position of the arm and the target from the state vector at the end of the reaching movement. 
Matrices $Q_{T_j}$ and $R$ define the precision- and the control-dependent cost, respectively (for more details see~\cite{christopoulos2015biologically}). 
The model implements a winner-take-all strategy, where the neuron with the highest activity level wins the competition and determines the reach direction. 
At the point when the activity of a neuronal group surpasses the action initiation threshold, a decision is determined and the associated reaching movement is executed~\cite{christopoulos2015biologically,zhong2022neurocomputational,zhong2023computational}.

\begin{figure}[!t]
\vspace{6pt}
    \centering
    \includegraphics[trim={5cm 2.8cm 6cm 2cm},clip, height=0.21\textheight]{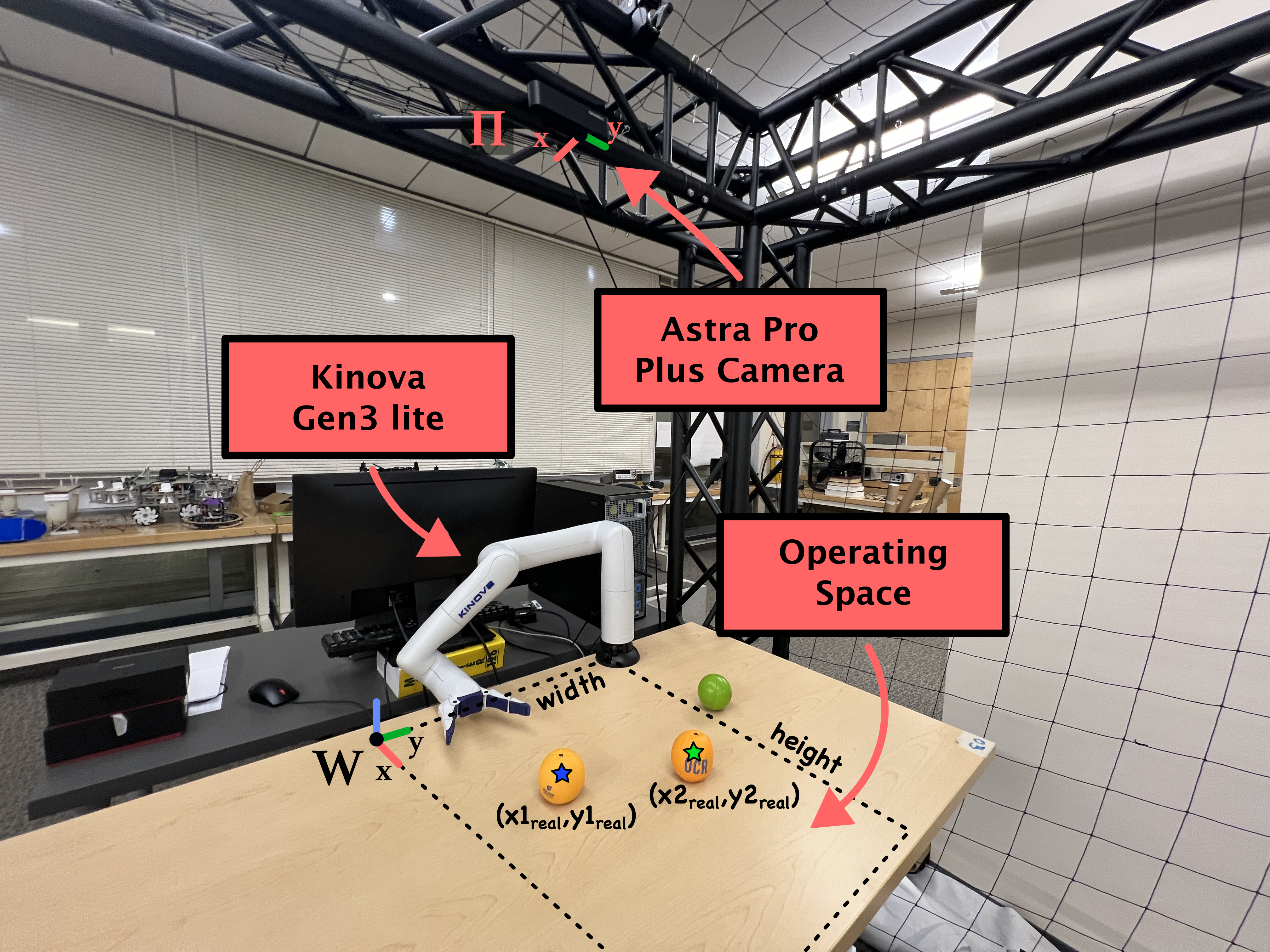}
    \caption{The robotic arm setup and sensing configuration. A camera provides bird-eye view feedback to the controller. The ball objects on the table serve as the main target beacons.}
    \label{fig:robot_setup}
    \vspace{-18pt}
\end{figure}

\subsection{Experimental Setup}

To evaluate our framework we employed the Kinova Gen3-lite robotic manipulator, fixed-mounted on a tabletop, to perform direct and dynamically-altering reaching scenarios. The arm is set up in a planar configuration, with its end-effector moving in the horizontal plane and on the same level as its base. 
The Kinova Gen3-lite has a reaching range of 76~$cm$ and a maximum speed of 25~$cm/sec$. 
The selected operating space on the table is 52 $\times$ 47~$cm^2$ in size and it was used for all our experiments. 
We used the manufacturer's software (Kinova Kortex) to send motor commands to the onboard controller and actuators. 

An Astra Pro Plus RGB camera ($30~fps$ live camera feed with 1080$p$ resolution via a USB 2.0 connection) was used for scene understanding and visual-based control. 
The camera was fixed above the robotic arm to provide a birds-eye view of the configuration space and detect the desired objects. 
The arm and camera were connected to a desktop computer, where all information was aggregated for the controller to compute the next robot action. 
We used plastic balls (7.5~$cm$ diameter) to denote different targets and associated actions. 
Orange and green balls act as visual beacons for target-reaching and stopping actions, respectively. 
No other objects were placed in the configuration space. 

Figure~\ref{fig:robot_setup} depicts the setup. 
We developed an application based on the Robot Operating System (ROS) and software packages in C++ (for robot control) and MATLAB (for NeuCF implementation). 
The initial end-effector position was set the same across all experiments, and without loss of generality, it was selected as the origin of the world frame. 
Desired end-effector positions, as well as, any intermediate positions computed by the controller were also given in the world frame, and trajectory generation ensued to determine the appropriate low-level robot commands (i.e. at the joint level). 
The latter was computed by directly invoking the robot arm manufacturer's inverse kinematics solver.

\subsection{Scene Understanding}



First, an affine transformation was used to compute beacon positions in space based on image plane measurements, i.e. 
\begin{equation}
\begin{bmatrix}
x' \\ y'
\end{bmatrix} = M \cdot \begin{bmatrix}
x \\ y \\ 1
\end{bmatrix} \Longleftrightarrow
\begin{bmatrix}
x' \\ y'
\end{bmatrix} = \begin{bmatrix}
a_{00} & a_{01} & b_{00} \\
a_{10} & a_{11} & b_{10} \\
\end{bmatrix} \cdot \begin{bmatrix}
x \\ y \\ 1
\end{bmatrix}\;,\label{eq:affinetransformation}
\end{equation}

\noindent where $M$ is the transformation matrix between camera frame $\Pi$ points ($x,y$) and orthographic frame $\mathcal{O}$ points ($x',y'$).
Matrix $M$ is initialized by using any of the three table corner coordinates in the $\Pi$ frame, and must be recomputed (i.e. perform extrinsic calibration) if the table position changes.


To acquire the beacons' real position in space, we extracted and removed the table background, to exclude false positives during localization. 
An HSV color model transformation was applied on the image plane $\mathcal{O}$ to find the corresponding beacon colors' given their hue value. 
Binary masking was then applied to obtain the orange and green areas on the image plane. 
Areas that were less than 15\% of the image plane size were excluded from detection. 
The coordinates of the beacons were then mapped to the real coordinates $(x_{real},y_{real})\in\mathcal{W}$ as
$x_{real} = (x'/x'_{max}) \cdot width$ and 
$y_{real} = (y'/y'_{max}) \cdot height$, where variables $width$ and $height$ correspond to the workspace planar boundaries expressed in frame $\mathcal{W}$ (Fig.~\ref{fig:robot_setup}) and $(x'_{max},y'_{max})$ are their equivalent representation in frame $\mathcal{O}$. 
In this way, pixel coordinates $(x', y')\in \mathcal{O}$ of a detected beacon are mapped into $\mathcal{W}$ coordinates, $(x_{real},y_{real})$ expressed in $cm$. The origin was set at the bottom-right corner of the table.  


\begin{figure*}
\vspace{4pt}
    \centering
    \begin{subfigure}[b]{0.19\textwidth}
        \includegraphics[width=\textwidth]{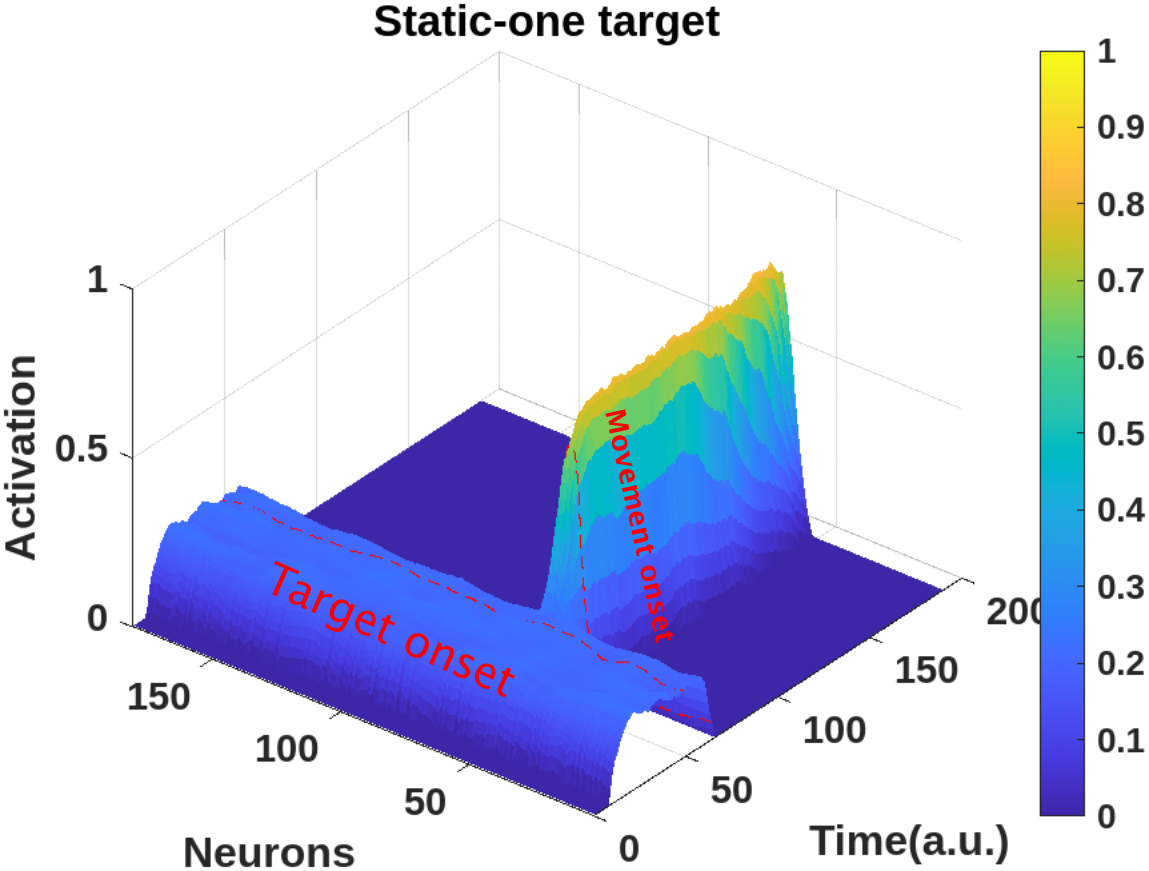}
        \caption{}
    \end{subfigure}
    \begin{subfigure}[b]{0.18\textwidth}
        \includegraphics[width=\textwidth]{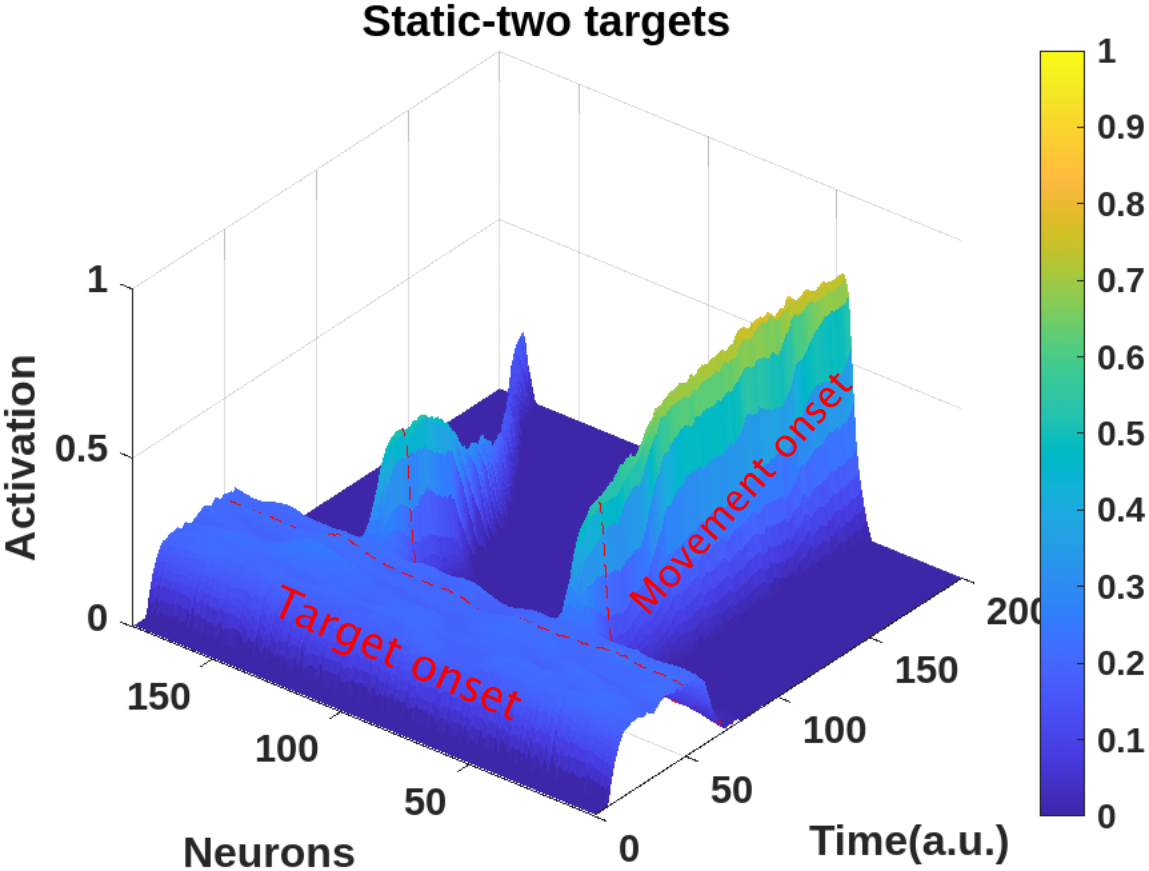}
        \caption{}
    \end{subfigure}
    \begin{subfigure}[b]{0.18\textwidth}
        \includegraphics[width=\textwidth]{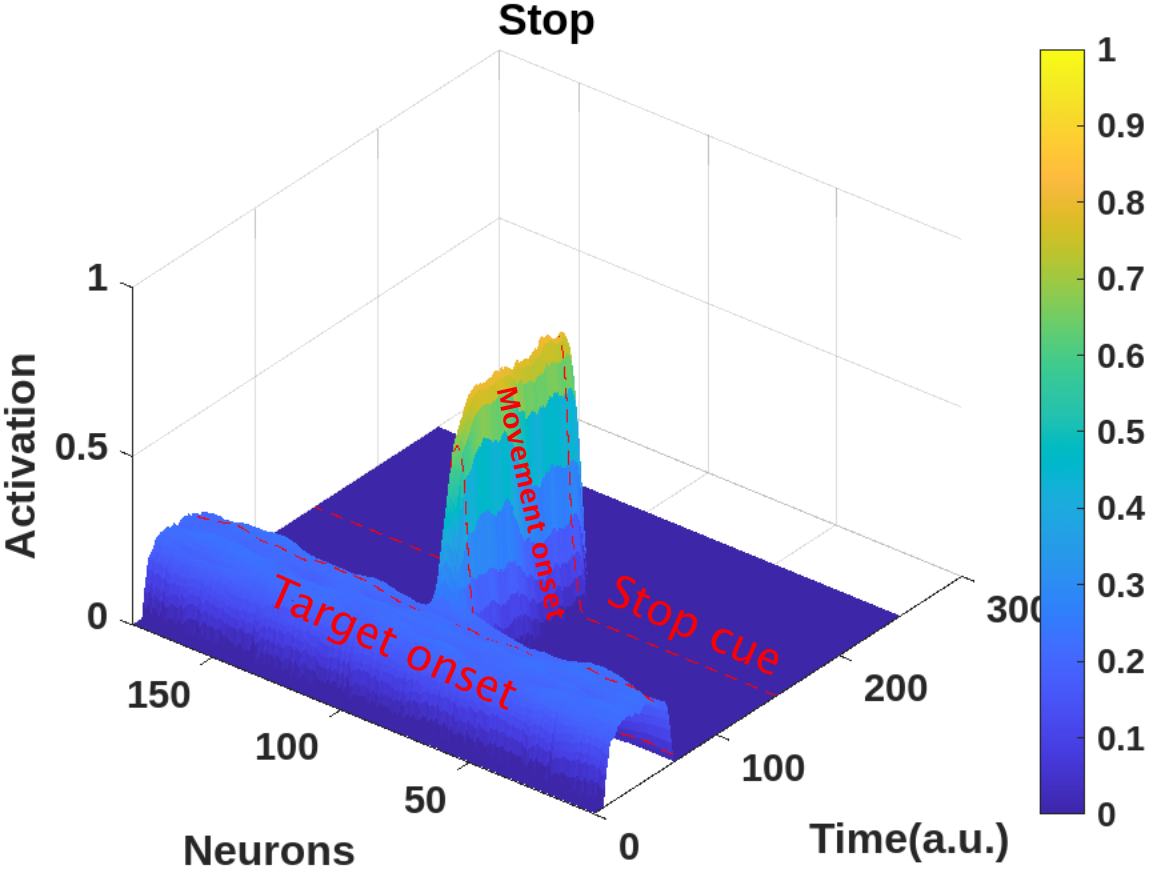}
        \caption{}
    \end{subfigure}
    \begin{subfigure}[b]{0.18\textwidth}
        \includegraphics[width=\textwidth]{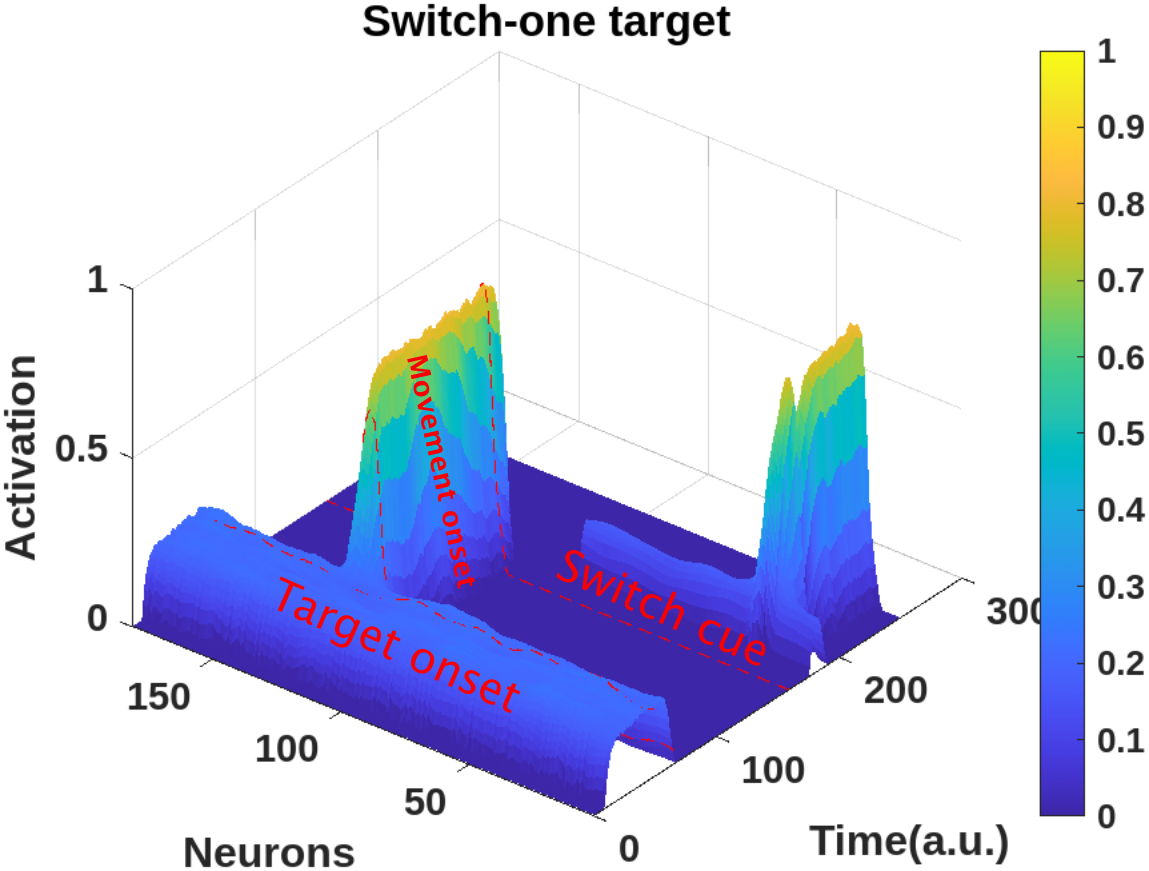}
        \caption{}
    \end{subfigure}
        \begin{subfigure}[b]{0.18\textwidth}
        \includegraphics[width=\textwidth]{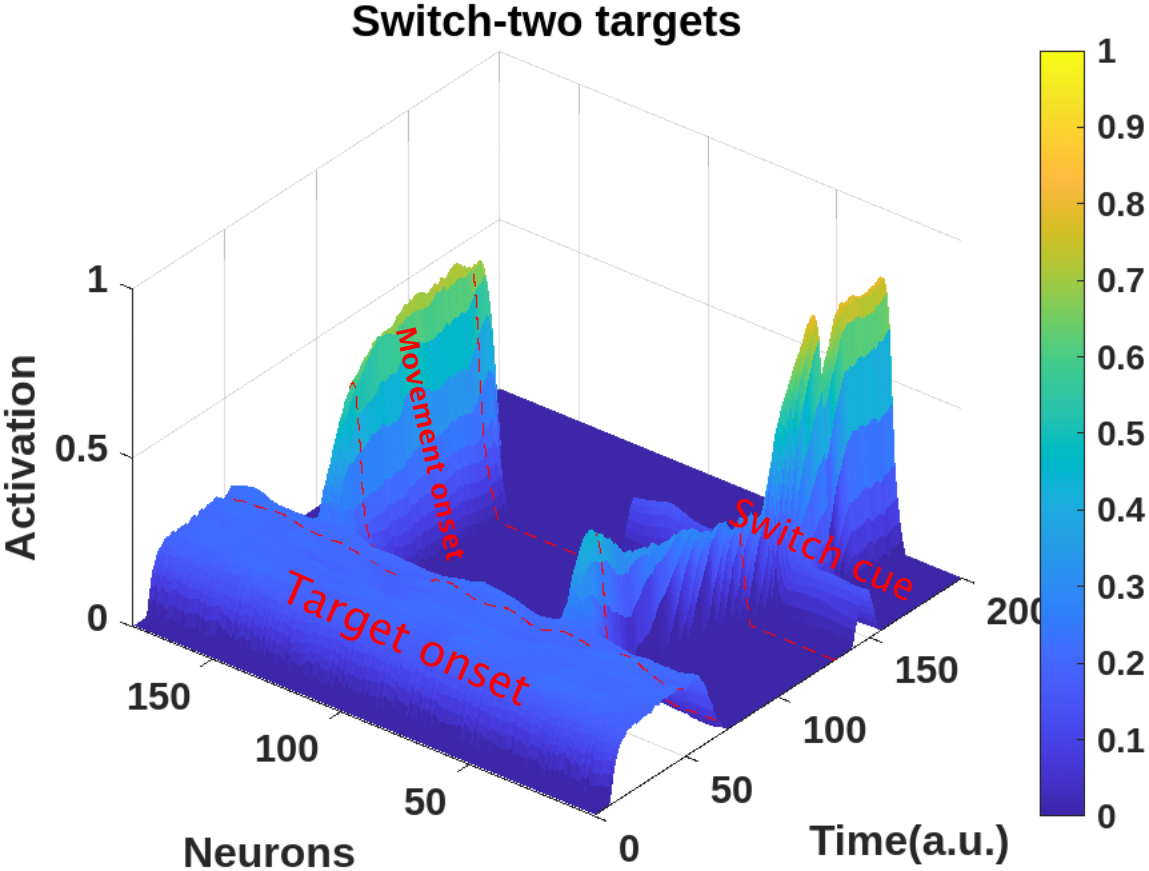}
        \caption{}
    \end{subfigure}
    \vspace{-5pt}
    \caption{Activity changes of the 181 neurons of the reach planning field during experiments. Each panel includes the instants the target appears (target onset), movement initiation (movement onset), and when the stop/switch cue appears (stop cue/switch cue). In case (d), only one beacon is available initially, and then we switch it with a new one, while in (e) both beacons appear initially and one beacon is then removed.}
    \label{fig:reachingplanningfields}
    \vspace{-18pt}
\end{figure*}

\subsection{Beacon State Description}

To describe the status of a beacon we define the 4-tuple $b_t:=\{visibility, (x',y'),(x_{real},y_{real}), t_{vis}\}$. 
Variable $visibility\in\{\text{``stationary," ``moving," ``disappeared"}\}$ indicates the appearance status of the corresponding beacon, while $t_{vis}$ holds the last timestamp that the beacon was detected. 
%
All beacon status information is updated in real-time with each received RGB frame. 
In the case that a beacon disappears we have set a 3 $sec$ delay after $t_{vis}$ to change its status from ``stationary'' or ``moving'' to ``disappeared.'' 
A beacon is considered as ``moving'' if its position in the x- and y-axis has changed for more than 5$\%$ with respect to $x'_{max}$ and $y'_{max}$, respectively. 
All this information is transmitted to the decision and contextual information modules of the NeuCF controller, to provide the current status of the appeared beacons during the reaching scenario. The state of each beacon is important during our experiments since the NeuCF system can adjust the reaching action in real time depending on observed beacon placement. 





\subsection{Polynomial Trajectory Generator}

A cubic polynomial trajectory generator was used to determine the time scaling function, $s(t)$, of the path given a desired goal on $\mathcal{W}$ frame and a terminal time $T$. 
This is computed as
$s(t) = a_0 + a_1t + a_2t^2 + a_3t^3$, with $t\in[0,T]$. 
Instants $s(t=0)$ and $s(t=T)$ denote the initial and goal position of the end-effector. 
The end-effector also needs to start and end at rest, hence $\dot{s}(T)=0$ and $\dot{s}(0)=0$. These boundary conditions help set and solve a linear program to compute the coefficients $a_i,i\in[0,3]$, yielding $a_0 = 0 \text{, } a_1 = 0\text{, } a_2 = \frac{3}{T^2}\text{, and } a_3 = -\frac{2}{T^3}$. 
Since we operate in an open area, trajectory generation along the x- and y- axes can be decoupled, and the above process is thus applied twice. 
We can obtain the path from the cubic polynomial trajectory generator by providing the total operation time (adjusted to the NeuCF controller) and the desired goal position.



\vspace{-3pt}
\section{Experiments, Results, and Discussion}\label{seq:experiments}

To evaluate the developed system we conducted five different experiments of reaching scenarios by using one or two target beacons and the stopping beacon. 
We considered three settings: 1) $static$ selection of reaching a stationary existing beacon, 2) the $switch$ beacon scenario when the goal target is re-positioned while approaching the target beacon, and 3) the sudden $stop$ scenario while reaching for a target beacon. 
In all cases, we also evaluated the polynomial trajectory generator baseline and compared its output with our system's generated path smoothness and reaching accuracy. 

For each experiment, we employed three different beacon setups and repeated each experiment three times. 
Each setup was tested and repeated the same amount of times for both the NeuCF and polynomial trajectory generation methods. 
For fairness, the polynomial trajectory generator was set to execute the trajectory at the same operating time as the NeuCF controller. 
We evaluated our proposed system using path smoothness, reaching goal positional accuracy, generated path length, and higher derivatives of x- and y-axis motion to assess the consistency of acceleration and jerk during motion. 
We also present the neuronal activation surface plots of the reaching planning field from each case, to demonstrate the activity changes that occurred.

\subsection{Static Beacon Selection}

\begin{table}[!t]
\vspace{4pt}
\centering
\caption{Experimental Results for $static$ Cases}
\vspace{-3pt}
\begin{tabular}
{c@{\hspace{0.5em}}c@{\hspace{0.5em}}c@{\hspace{0.5em}}|c@{\hspace{0.5em}}c@{\hspace{0.5em}}}
\toprule
         Scenario    & \multicolumn{2}{c}{$static\_1$}                                      & \multicolumn{2}{c}{$static\_2$}                        \vspace{0.1cm}               \\ \toprule 
          Controller   & NeuCF   & Polynomial & NeuCF   & Polynomial \\ \midrule 
X Error $(cm)$ & 0.32$\pm$0.02         & 0.20$\pm$0.01          & 0.20$\pm$0.12         & 0.12$\pm$0.04          \\
Y Error $(cm)$ & 0.39$\pm$0.28         & 0.74$\pm$0.12          & 0.87$\pm$0.14         & 0.77$\pm$0.01          \\
Path Length $(cm)$ & 31.02  & 30.92      & 28.22   & 28.83    \\
Straightness $r^2$ & 0.9994 & 1    & 0.9985 & 0.9999   \\
\bottomrule
\end{tabular}\label{tab:stationarychoices}
\vspace{-5pt}
\end{table}
 
\begin{figure}[!t]
\vspace{-9pt}
    \centering
    \begin{subfigure}[b]{0.23\textwidth}
        \includegraphics[width=\textwidth]{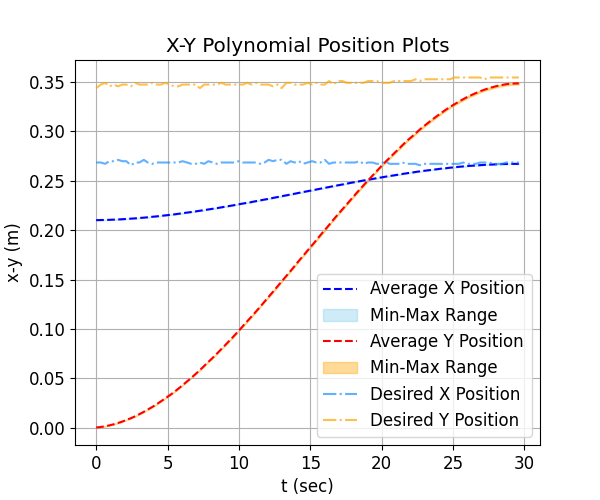}
        \vspace{-15pt}
        \caption{}
    \end{subfigure}
    \vspace{-1pt}
    \begin{subfigure}[b]{0.23\textwidth}
        \includegraphics[width=\textwidth]{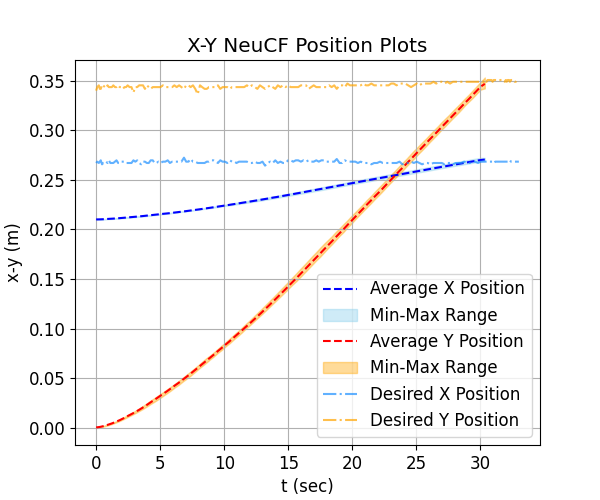}
        \vspace{-15pt}
        \caption{}
    \end{subfigure}
    \vspace{-1pt}
    \begin{subfigure}[b]{0.23\textwidth}
        \includegraphics[width=\textwidth]{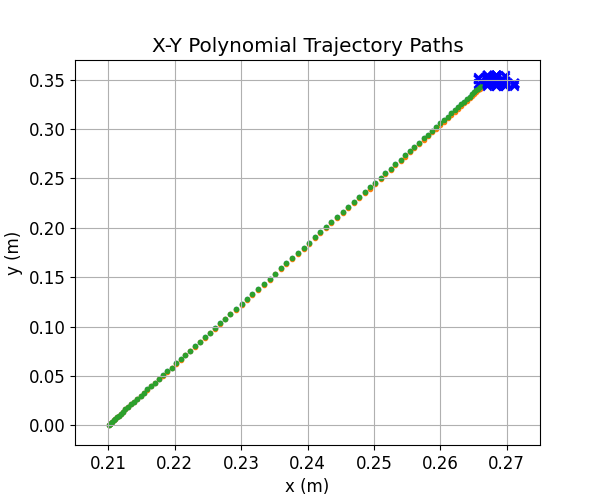}
        \vspace{-15pt}
        \caption{}
    \end{subfigure}
    \begin{subfigure}[b]{0.23\textwidth}
        \includegraphics[width=\textwidth]{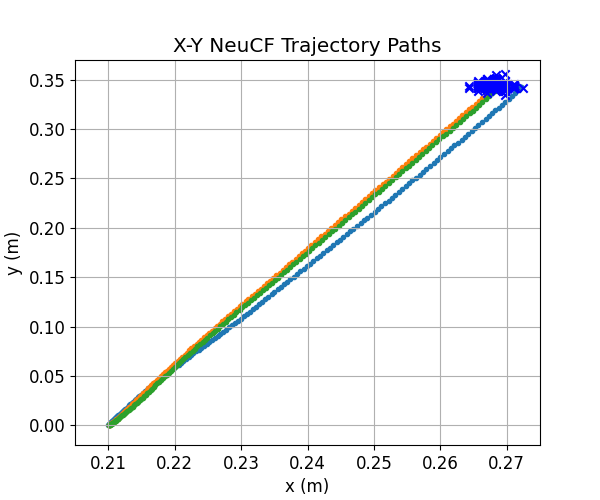}
        \vspace{-15pt}
        \caption{}
    \end{subfigure}
    \vspace{-3pt}
    \caption{Resulting trajectories given a selected stationary beacon at (27, 35) $cm$ in the $static\_1$ scenario. Panels (a) and (b) show the evolution of the x- and y-axis end-effector positions for the Polynomial and NeuCF controllers, respectively; top-down views are shown in panels (c) and (d).}
    \label{fig:xypositioning}
    \vspace{-18pt}
\end{figure}

We first evaluated the generation and execution of direct paths toward a selected target (out of possibly two) that remains static throughout the experiment. 
Table~\ref{tab:stationarychoices} contains the obtained results. 
It becomes evident upon perusal that the NeuCF-based controller performs on par with the baseline polynomial trajectory generator. 
Goal-reaching accuracy was high, whereby obtained trajectories have less than 0.87~$cm$ of absolute positioning error in both the x- and y-axis, on average. 
The one-standard deviation of the positioning error of the end-effector was less than 0.14~$cm$ in the examined cases. 
An example is depicted in Fig.~\ref{fig:xypositioning}, showing the NeuCF-based controller can generate a more direct path to the target.

To evaluate trajectory smoothness, we performed linear regression on the generated x-y trajectory and calculated the coefficient of determination $r^2$ (whereby a value closer to 1 corresponds to more straight and direct trajectories). 
NeuCF generates trajectories with $r^2>0.99$ (Table~\ref{tab:stationarychoices}). 
Attained path lengths are almost equal in length to the ones generated by the direct polynomial trajectory system (over 96\% match). 
Figures~\ref{fig:reachingplanningfields}a and~\ref{fig:reachingplanningfields}b show the neuronal activations from initiation of movement toward the selected target. The average total runtime of the experiments was 27.5~$sec$.



\subsection{Reaching Interruption}

\begin{table}[t!]
\vspace{6pt}
\centering
\caption{Experimental Results for $stop$ Cases}
\vspace{-3pt}
\begin{tabular}{ccccc}
\toprule
        Scenario     & \multicolumn{2}{c}{$stop$}                                                     \vspace{0.1cm}  \\ \toprule
           Controller  & NeuCF             & Polynomial                \\ \midrule 
Path Length ($cm$)  & 16.15           & 15.33           \\
Straightness $r^2$ & 0.9983            & 0.9999            \\
Acceleration ($cm/sec^2$) & 0.051$\pm$0.25 & 0.052$\pm$0.07 \\
Jerk  ($cm/sec^3$)       & 0.173$\pm$2.55 & 0.009$\pm$0.18 \\ \bottomrule
\end{tabular} \label{tab:stop}
\vspace{-11pt}
\end{table}

\begin{figure}[t!]
    \centering
    \begin{subfigure}[b]{0.23\textwidth}
        \includegraphics[width=\textwidth]{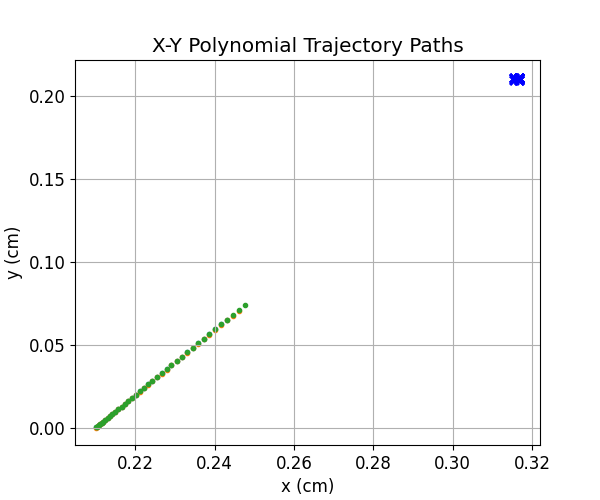}
        \vspace{-17pt}
        \caption{}
    \end{subfigure}
    \begin{subfigure}[b]{0.23\textwidth}
        \includegraphics[width=\textwidth]{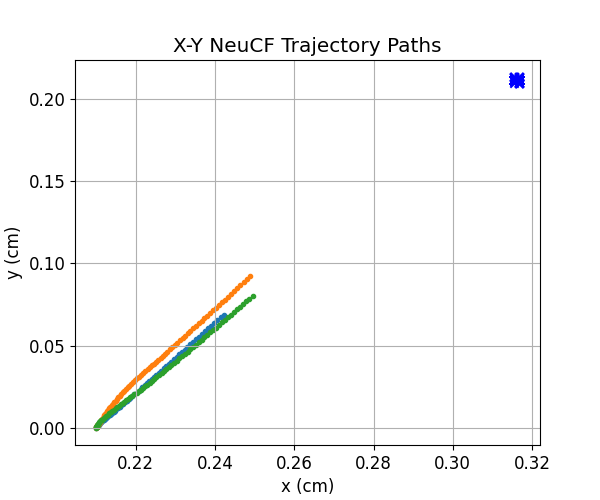}
        \vspace{-17pt}
        \caption{}
    \end{subfigure}
    \vspace{-5pt}
    \caption{Generated trajectories by (a) the polynomial and (b) NeuCF controllers during a $stop$ experiment. When the green stop beacon appears, both controllers receive an interruption signal and can stop the reaching action successfully.}
    \label{fig:xypositioningstop}
    \vspace{-18pt}
\end{figure}

We then moved on with testing the ability to interrupt the execution of an ongoing reaching action when a stopping beacon appears in the workspace. This can be applied in either $static$ scenarios where the beacons are stationary or during the $switch$ scenarios (discussed next) when the robot is trying to reach a new target beacon because of a change in the environment. 
Obtained results are shown in Table~\ref{tab:stop}. 
The system can afford successful interruption of the reaching action when the stopping beacon appears. 
Trajectories remain smooth until the end-effector comes to a full stop. 
The stopping signal does not cause any instability in the last part of the arm's trajectory (Fig.~\ref{fig:xypositioningstop}). The reported path acceleration and jerk attained relatively low values, although not as smooth compared with the polynomial controller. 
Acceleration variation was less than 27\% compared with the polynomial controller, with the maximum one-standard deviation at $0.07~cm/sec^2$. Variation in jerk in both x- and y-axis was less than $0.30~cm/sec^3$, with the one-standard deviation at $2.6~cm/sec^3$. Figure~\ref{fig:reachingplanningfields}c shows the interrupted neuronal activations after the stopping cue has been received. 

\subsection{Switching Beacon Choice}


In the $switch$ cases we test the dynamic adaptability of our system when the beacons are re-positioned or removed in the middle of executing a reaching action. 
We consider two cases. 
In the first case ($switch\_1$) one beacon exists in the field which the NeuCF controller selects and generates a path toward it in real-time. 
During execution, the beacon is removed and a new target beacon is placed at a new position for the robot to reach. 
In the second case ($switch\_2$) two beacons appear initially and the arm starts reaching toward one. 
Then, we remove the selected target to force the controller to switch its target. 
During all the experiments in each case, the beacons were removed and replaced at the same time instant and were placed in the same position. We conducted three different beacon placement setups for both $switch\_1$ and $switch\_2$ cases. 
We set a time limit at 36~$secs$.

\begin{table}[t!]
\vspace{6pt}
\centering
\caption{Experimental Results for $switch$ Cases}
\vspace{-3pt}
\begin{tabular}{c@{\hspace{0.5em}}c@{\hspace{0.5em}}c@{\hspace{0.5em}}|c@{\hspace{0.5em}}c@{\hspace{0.5em}}}
\toprule 
Scenario             & \multicolumn{2}{c}{$switch\_1$}                                                   & \multicolumn{2}{c}{$switch\_2$}                                               \vspace{0.1cm}    \\ \toprule
        Controller     & NeuCF             & Polynomial        & NeuCF             & Polynomial        \\ \midrule
$2^{nd}$ Der. Variance           & 5.10e-9        & 1.59e-8        & 4.85e-9        & 3.66e-8        \\
Fractal Dimension      & -0.668        & -0.744        & -0.767        & -0.784        \\
Accel. ($cm/sec^2$) & 0.01$\pm$0.20 & 0.02$\pm$0.09 & 0.01$\pm$0.22 & 0.02$\pm$0.18 \\
Jerk  ($cm/sec^3$)           & 0.11$\pm$1.62 & 0.01$\pm$0.20 & 0.15$\pm$1.92 & 0.01$\pm$0.39 \\ \bottomrule
\end{tabular} \label{tab:switch}
\vspace{-13pt}
\end{table}

\begin{figure}[t!]
    \centering
    \begin{subfigure}[b]{0.23\textwidth}
        \includegraphics[width=\textwidth]{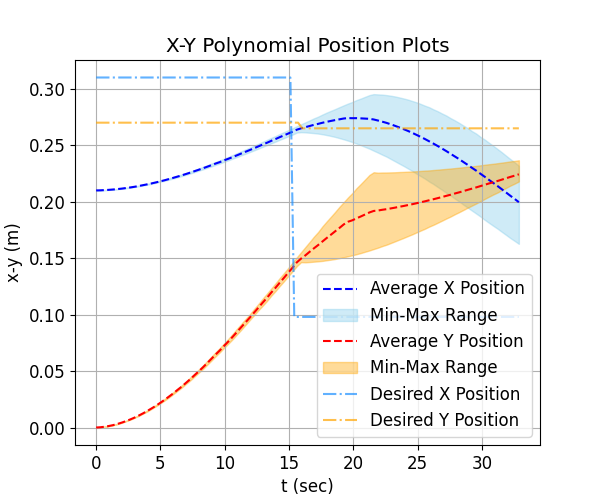}
        \vspace{-17pt}
        \caption{}
        \vspace{2pt}
    \end{subfigure}
    \begin{subfigure}[b]{0.23\textwidth}
        \includegraphics[width=\textwidth]{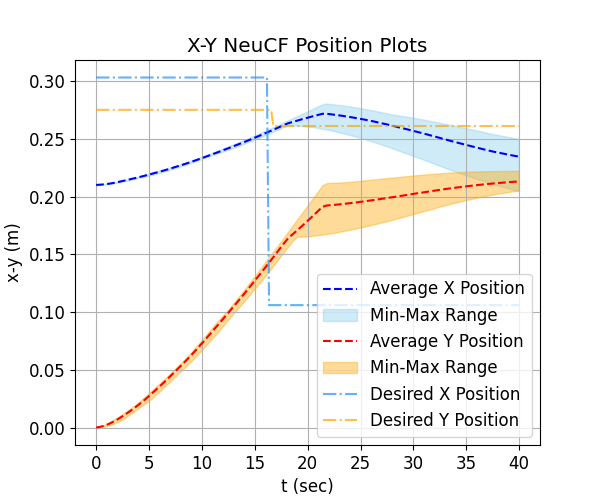}
        \vspace{-17pt}
        \caption{}
        \vspace{2pt}
    \end{subfigure}
    \begin{subfigure}[b]{0.23\textwidth}
        \includegraphics[trim={0cm 0cm 0cm 0.8cm},clip,width=\textwidth]{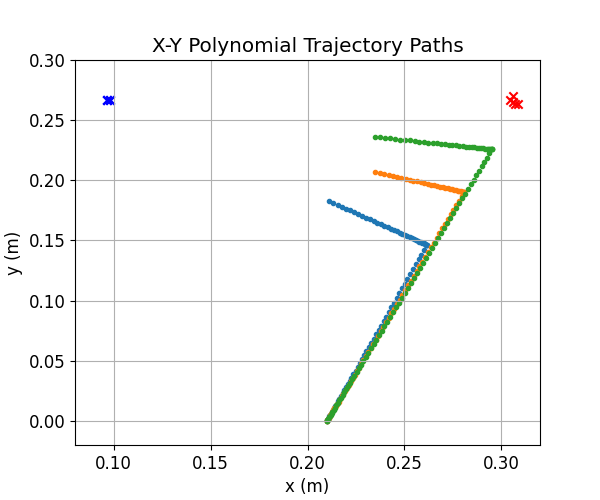}
        \vspace{-17pt}
        \caption{}
    \end{subfigure}
    \begin{subfigure}[b]{0.23\textwidth}
        \includegraphics[trim={0cm 0cm 0cm 0.8cm},clip,width=\textwidth]{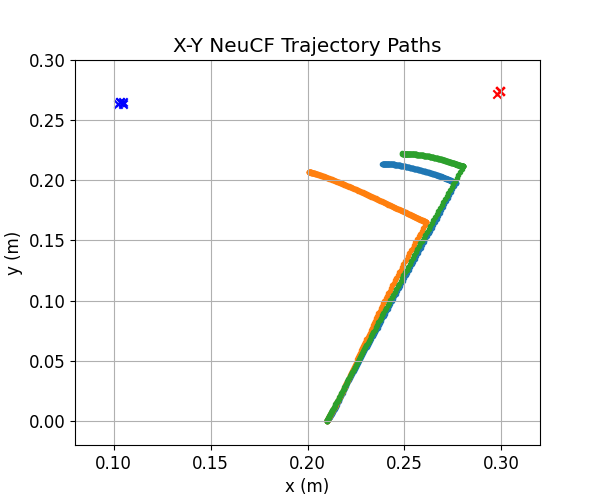}
        \vspace{-17pt}
        \caption{}
    \end{subfigure}
    \vspace{-6pt}
    \caption{Experiments during a $switch\_2$ case. Two target beacons are placed and the target at the top-right is selected (in red color). During the reaching action, we remove the target beacon and thus prompt the controllers to re-schedule and approach the other target at top-left (in blue color).}
    \label{fig:xypositioningswitch}
    \vspace{-18pt}
\end{figure}

Since the generated trajectories are more complex due to the on-demand target switch, we employ second derivative and fractal dimension analysis to evaluate the path smoothness. For the former, we numerically compute the second derivative of a trajectory and then measure its variance. In the latter, we analyze the generated plots into a fractal pattern of 50 different box scales and calculate the fractal dimension which reflects the path's spatial complexity. 

Results are presented in Table~\ref{tab:switch}. 
There is high linearity in the second derivative analysis providing smooth curves when transitioning to a new target. Scores are overall better compared to the polynomial-based trajectory (smaller positioning variance at $4.98\cdot10^{-9}$ compared with the $2.62\cdot10^{-8}$ of the cubic polynomial). 
Fractal analysis also shows the smaller complexity of the NeuCF-generated trajectories by scoring up to 11\% smaller fractal score. 
Figures~\ref{fig:xypositioningswitch}a and~\ref{fig:xypositioningswitch}b depict one of the $switch\_2$ cases showing both methods output when switching a target while  
Fig.~\ref{fig:xypositioningswitch}c and~\ref{fig:xypositioningswitch}d show the independent trials during the $switch\_2$ case. 
Obtained acceleration in both $switch$ cases was similar between the two controllers, but 
NeuCF had higher values and variance on path jerk because of the direct re-targeting. 
Figures~\ref{fig:reachingplanningfields}d and~\ref{fig:reachingplanningfields}e show the neuronal activations for switching NeuCF behavior to reach the available target.

\vspace{-3pt}
\section{Conclusions}

We demonstrated the potential of the NeuCF controller to perform accurate and dynamic environment-aware target reaching for robotic manipulation. 
Visual feedback from an off-body camera offers target and robot end-effector position information in real-time which is in turn used by the NeuCF controller to generate smooth reaching trajectories toward a selected target. 
Several tested cases, including static target reaching, action stopping, and switching to a different target based on changes in the environment at runtime, highlight that the NeuCF controller can afford dynamic real-time re-prioritization for robot reaching. 
Results showed its robustness in positioning accuracy and smooth trajectory generation and yielded similar results to a baseline polynomial trajectory generator. 
Future work will focus on 3D reaching. 
\vspace{-12pt}





\bibliographystyle{ieeetr}
\bibliography{ref}

\begin{thebibliography}{10}

\bibitem{li2018rlmanipulationhumanoid}
Z.~Li, T.~Zhao, F.~Chen, Y.~Hu, C.-Y. Su, and T.~Fukuda, ``Reinforcement learning of manipulation and grasping using dynamical movement primitives for a humanoidlike mobile manipulator,'' {\em IEEE/ASME Transactions on Mechatronics}, vol.~23, no.~1, pp.~121--131, 2018.

\bibitem{zimmermann2021spotminigrasp}
S.~Zimmermann, R.~Poranne, and S.~Coros, ``Go fetch! - dynamic grasps using {B}oston {D}ynamics spot with external robotic arm,'' in {\em IEEE International Conference on Robotics and Automation (ICRA)}, pp.~4488--4494, 2021.

\bibitem{shi2022multifinger}
L.~Shi, C.~Mucchiani, and K.~Karydis, ``Online modeling and control of soft multi-fingered grippers via {K}oopman operator theory,'' in {\em IEEE 18th International Conference on Automation Science and Engineering (CASE)}, pp.~1946--1952, 2022.

\bibitem{mandlekar2020irisofflinemanipulationdata}
A.~Mandlekar, F.~Ramos, B.~Boots, S.~Savarese, L.~Fei-Fei, A.~Garg, and D.~Fox, ``Iris: Implicit reinforcement without interaction at scale for learning control from offline robot manipulation data,'' in {\em IEEE International Conference on Robotics and Automation (ICRA)}, pp.~4414--4420, 2020.

\bibitem{katyara2021humaninspiredrobotmanipulation}
S.~Katyara, F.~Ficuciello, F.~Chen, B.~Siciliano, and D.~G. Caldwell, ``Vision based adaptation to kernelized synergies for human inspired robotic manipulation,'' in {\em IEEE International Conference on Robotics and Automation (ICRA)}, pp.~6491--6497, 2021.

\bibitem{hueang2022prostheticbiosignals}
J.~Huang, G.~Li, H.~Su, and Z.~Li, ``Development and continuous control of an intelligent upper-limb neuroprosthesis for reach and grasp motions using biological signals,'' {\em IEEE Transactions on Systems, Man, and Cybernetics: Systems}, vol.~52, no.~6, pp.~3431--3441, 2022.

\bibitem{ciocarlie2009hand}
M.~T. Ciocarlie and P.~K. Allen, ``Hand posture subspaces for dexterous robotic grasping,'' {\em The International Journal of Robotics Research}, vol.~28, no.~7, pp.~851--867, 2009.

\bibitem{erlhagen2006DNFcognitiverobot}
W.~Erlhagen and E.~Bicho, ``The dynamic neural field approach to cognitive robotics,'' {\em Journal of Neural Engineering}, vol.~3, p.~R36, jun 2006.

\bibitem{hersch2006model}
M.~Hersch and A.~G. Billard, ``A model for imitating human reaching movements,'' in {\em Proceedings of the 1st ACM SIGCHI/SIGART conference on Human-robot interaction}, pp.~341--342, 2006.

\bibitem{hyondong2017bioinspiredmultirobots}
H.~Oh, A.~{Ramezan Shirazi}, C.~Sun, and Y.~Jin, ``Bio-inspired self-organising multi-robot pattern formation: A review,'' {\em Robotics and Autonomous Systems}, vol.~91, pp.~83--100, 2017.

\bibitem{dayan2005neurosiencetheoretical}
P.~Dayan and L.~F. Abbott, {\em Theoretical neuroscience: computational and mathematical modeling of neural systems}.
\newblock MIT press, 2005.

\bibitem{steffen2020snn3dplanning}
L.~Steffen, R.~K.~d. Silva, S.~Ulbrich, J.~C.~V. Tieck, A.~Roennau, and R.~Dillmann, ``Networks of place cells for representing 3d environments and path planning,'' in {\em 8th IEEE RAS/EMBS International Conference for Biomedical Robotics and Biomechatronics (BioRob)}, pp.~1158--1165, 2020.

\bibitem{camilo2018ssnmotorreaching}
J.~C.~V. Tieck, L.~Steffen, J.~Kaiser, A.~Roennau, and R.~Dillmann, ``Controlling a robot arm for target reaching without planning using spiking neurons,'' in {\em IEEE 17th International Conference on Cognitive Informatics \& Cognitive Computing (ICCI* CC)}, pp.~111--116, 2018.

\bibitem{vasquez2019ssnrlreaching}
J.~C. Vasquez~Tieck, P.~Becker, J.~Kaiser, I.~Peric, M.~Akl, D.~Reichard, A.~Roennau, and R.~Dillmann, ``Learning target reaching motions with a robotic arm using brain-inspired dopamine modulated {STDP},'' in {\em IEEE 18th International Conference on Cognitive Informatics \& Cognitive Computing (ICCI*CC)}, pp.~54--61, 2019.

\bibitem{weiSNNflightcontroller_faulttolerant}
W.~Yu, N.~Yang, Z.~Wang, H.~C. Li, A.~Zhang, C.~Mu, and S.~H. Pun, ``Fault-tolerant attitude tracking control driven by spiking nns for unmanned aerial vehicles,'' {\em IEEE Transactions on Neural Networks and Learning Systems}, pp.~1--13, 2023.

\bibitem{tanaka2019reservoir}
G.~Tanaka, T.~Yamane, J.~B. H{\'e}roux, R.~Nakane, N.~Kanazawa, S.~Takeda, H.~Numata, D.~Nakano, and A.~Hirose, ``Recent advances in physical reservoir computing: A review,'' {\em Neural Networks}, vol.~115, pp.~100--123, 2019.

\bibitem{maass2002lsmbook}
W.~Maass, T.~Natschläger, and H.~Markram, ``Real-time computing without stable states: A new framework for neural computation based on perturbations,'' {\em Neural Computation}, vol.~14, no.~11, pp.~2531--2560, 2002.

\bibitem{alberto2017positioninglsm}
D.~Alberto~Sala, V.~João~Brusamarello, R.~de~Azambuja, and A.~Cangelosi, ``Positioning control on a collaborative robot by sensor fusion with liquid state machines,'' in {\em IEEE International Instrumentation and Measurement Technology Conference (I2MTC)}, pp.~1--6, 2017.

\bibitem{de2017robotarmlsm}
R.~de~Azambuja, F.~B. Klein, S.~V. Adams, M.~F. Stoelen, and A.~Cangelosi, ``Short-term plasticity in a liquid state machine biomimetic robot arm controller,'' in {\em International Joint Conference on Neural Networks (IJCNN)}, pp.~3399--3408, 2017.

\bibitem{Sandamirskaya_2014}
S.~Y, ``Dynamic neural fields as a step toward cognitive neuromorphic architectures,'' {\em Frontiers in Neuroscience}, vol.~7, no.~276, 2014.

\bibitem{faubel2008learning}
C.~Faubel and G.~Sch{\"o}ner, ``Learning to recognize objects on the fly: a neurally based dynamic field approach,'' {\em Neural Networks}, vol.~21, no.~4, pp.~562--576, 2008.

\bibitem{katerishich2023dnfautonomousnavigation}
M.~Katerishich, M.~Kurenkov, S.~Karaf, A.~Nenashev, and D.~Tsetserukou, ``Dnfomp: Dynamic neural field optimal motion planner for navigation of autonomous robots in cluttered environment,'' in {\em IEEE International Conference on Systems, Man, and Cybernetics (SMC)}, pp.~1984--1989, 2023.

\bibitem{cunha2020humanrobotaction}
A.~Cunha, F.~Ferreira, E.~Sousa, L.~Louro, P.~Vicente, S.~Monteiro, W.~Erlhagen, and E.~Bicho, ``Towards collaborative robots as intelligent co-workers in human-robot joint tasks: what to do and who does it?,'' in {\em 52th International Symposium on Robotics}, pp.~1--8, VDE, 2020.

\bibitem{malheiro2017frameworkDNFvrep}
T.~Malheiro, E.~Bicho, T.~Machado, L.~Louro, S.~Monteiro, P.~Vicente, and W.~Erlhagen, ``A software framework for the implementation of dynamic neural field control architectures for human-robot interaction,'' in {\em IEEE International Conference on Autonomous Robot Systems and Competitions (ICARSC)}, pp.~146--152, 2017.

\bibitem{erlhagen2002dynamic}
W.~Erlhagen and G.~Sch{\"o}ner, ``Dynamic field theory of movement preparation.,'' {\em Psychological review}, vol.~109, no.~3, p.~545, 2002.

\bibitem{ferreira2021rapidlearningdnf}
F.~Ferreira, W.~Wojtak, E.~Sousa, L.~Louro, E.~Bicho, and W.~Erlhagen, ``Rapid learning of complex sequences with time constraints: A dynamic neural field model,'' {\em IEEE Transactions on Cognitive and Developmental Systems}, vol.~13, no.~4, pp.~853--864, 2021.

\bibitem{strauss2015choice}
S.~Strauss, P.~J. Woodgate, S.~A. Sami, and D.~Heinke, ``Choice reaching with a lego arm robot corlego): the motor system guides visual attention to movement-relevant information,'' {\em Neural Networks}, vol.~72, pp.~3--12, 2015.

\bibitem{Knips_2017}
G.~Knips, S.~Zibner, H.~Reimann, and G.~Schöner, ``A neural dynamics architecture for grasping that integrates perception and movement generation and enables on-line updating,'' {\em Frontiers in Neurorobotics}, vol.~11, no.~9, 2017.

\bibitem{christopoulos2015probability}
V.~Christopoulos and S.~Paul, ``Dynamic integration of value information into a common probability currency as a theory for flexible decision making,'' {\em PLOS Computational Biology}, vol.~11, no.~9, p.~e1004402, 2015.

\bibitem{christopoulos2015biologically}
V.~Christopoulos, J.~Bonaiuto, and R.~A. Andersen, ``A biologically plausible computational theory for value integration and action selection in decisions with competing alternatives,'' {\em PLOS Computational Biology}, vol.~11, no.~3, p.~e1004104, 2015.

\bibitem{zhong2022neurocomputational}
S.~Zhong, J.~W. Choi, N.~G. Hashoush, D.~Babayan, M.~Malekmohammadi, N.~Pouratian, and V.~Christopoulos, ``A neurocomputational theory of action regulation predicts motor behavior in neurotypical individuals and patients with {P}arkinson’s disease,'' {\em PLOS Computational Biology}, vol.~18, no.~11, p.~e1010111, 2022.

\bibitem{zhong2023computational}
S.~Zhong, N.~Pouratian, and V.~Christopoulos, ``Computational mechanism underlying switching of motor plans,'' {\em bioRxiv}, 2023.

\end{thebibliography}


\end{document}